\documentclass[Conference]{IEEEtran}
\usepackage{graphicx}
\usepackage{subfigure}
\usepackage{multirow}
\usepackage{multicol}
\usepackage{stfloats}
\usepackage{cite}
\usepackage{amsmath,amssymb,amsfonts}
\usepackage{algorithmic}
\usepackage{textcomp}
\usepackage{xcolor}
\usepackage[justification=centering]{caption}
\begin{document}  
	
\title{Channel Pruned YOLOv5-based Deep Learning Approach for Rapid and Accurate Outdoor Obstacles Detection}


\author{Zeqian~Li,~\IEEEmembership{Student Member,~IEEE,}
	Yuwei~Wang,~\IEEEmembership{Student Member,~IEEE,}
	Kexun~Chen,~\IEEEmembership{Student Member,~IEEE,}
	Zhibin~Yu,~\IEEEmembership{Senior Member,~IEEE,}
}
	
	\maketitle   
	
	\begin{abstract}
		One-stage algorithm have been widely used in target detection systems that need to be trained with massive data. Most of them perform well both in real-time and accuracy. However, due to their convolutional structure, they need more computing power and greater memory consumption. Hence, we applied pruning strategy to target detection networks to reduce the number of parameters and the size of model. To demonstrate the practicality of the pruning method, we select the YOLOv5 model for experiments and provide a data set of outdoor obstacles to show the effect of model. In this specific data set, in the best circumstances, the volume of the network model is reduced by 49.7\% compared with the original model, and the reasoning time is reduced by 52.5\%. Meanwhile, it also uses data processing methods to compensate for the drop in accuracy caused by pruning. 
	\end{abstract}
	
	
	\section{INTRODUCTION}
	\IEEEPARstart{A} great number of CNN-based object detectors have been widely used in obstacles avoidance systems. For instance, inspecting for obstacle and pedestrians along roads through surveillance camera, whereas the quantity and variety of obstacle in a certain unit could be obtained through one target detection system. Object detector applications in real world are mainly affected by run-time memory, model size and number of computing operations, yet only some parts of them satisfy the above three demands. The state-of-the-arts detectors, including multiple stage R-CNN \cite{ref1}, series like R-FCN \cite{ref2} or single stage SSD \cite{ref3}, YOLO \cite{ref4} and many other extended works \cite{ref5}. VGGNet \cite{ref6}, DenseNet \cite{ref7}, CSPNet \cite{ref8}, MobileNet \cite{ref9, ref10}, or ShuffleNet \cite{ref11} could be the backbone for detectors running on CPU or GPU platforms. When compared to other state-of-the-art object detectors, the YOLOv5 operates in real-time and requires less computing. As a result, in our experiments, we use YOLOv5-based models.
	
	The major purpose of this research is to find the YOLOv5-based model that is both real-time and precise, Which techniques change only the training strategy or the accuracy of the network can be improved by increasing the training cost. The following is a summary of our contributions:
	
	We furnish a complex data set for model's training, which is gathered at various distances by an unmanned aerial vehicle (UAV), and it includes three meteorological conditions: sunny, cloudy, and foggy, as well contents the concepts of dense data set. Moreover, we compare the effects of several YOLOv5-based models on the full test set, including the time it takes to detect a single image, the mAP value, and other significant metrics, also we verified the effect of data pre-processing methods during detector training and determined which method can significantly improve mAP values.
	
	The rest of the paper is organized as follows. First, we briefly describe the features of the YOLOv3-based model. Secondly, the proposed model is detailed. Third, the provided dataset and experimental results are benchmarked. Finally, we conclude this work.

	\section{THE PROPOSED MODEL}
	In this section, we present the improvements on YOLOv5-based models', and then recommend the implementation process of YOLOv5-prune. 
	
	\subsection{YOLO-based Models}
	As a single-stage series network, YOLOv5 has higher accuracy and faster processing speed, but in real scenarios, it requires a large number of GPUs for high-volume training. Therefore, some network structures based on YOLOv5 have emerged.
	First, the traditional YOLOv3 model optimized with this method includes three different detection scales, increases the detection effect on small targets, modifies the loss function, predicts the confidence and class of targets by logistic regression, and uses multi-label classification instead of SoftMax classification. The complete network structure is shown in Figure 1. Second, YOLOv3 tiny aims to improve the detection speed. It differs from YOLOv3 in that it reduces the number of convolutional structures and fits only two detection layers, which significantly shortens the inference time and leads to a significant decrease in accuracy.
	
	\subsection{Pruning Process}
	Pruning refers to reducing the number of network parameters by removing channels that have less impact on the network after sparse training. Since applying YOLOv3 to real scenarios, we should prune while keeping the accuracy constant.
	
	In the design of YOLOv5, each convolutional structure consists of three network layers: the convolutional layer, the BN layer, and the activation layer, as shown in Figure 1, which are called CBRs. the BN layer serves to normalize the output of the previous convolutional layer, and this normalization is carried out throughout the neural network. On the one hand, this normalization forces to change the input to a relatively normal distribution with mean 0 and variance 1, so that the output of the BN layer can be easily used in the next Relu layer to avoid gradient vanishing. On the other hand, the fourth input of BN associates all samples in a small batch together, so that the network no longer generates deterministic values for a given training example \cite{ref12, ref13}. The distribution of the BN layer may not be greatly affected during the learning process of the entire BN layer since it may not be greatly affected during the learning process of the entire BN layer. Therefore, we choose to prune the BN layer of YOLOv5.
	
	In this article, YOLOv3-prune will be divided into the following two steps: 
	
	\subsubsection{Sparse training is performed on YOLOv5}
	In YOLOv5, it is assumed that $C_{out}$ and $B_{out}$ represent the output of the previous convolutional layer and the output of the BN layer, respectively, and $B_{mini}$ denotes the current small batch, and the BN layer undergoes the following transformation: $$B_{out} = \varepsilon * \breve{B} + \theta; \breve{B} = \frac{C_{out} - \beta_{b_{mini}}}{\sqrt{\alpha^{2}_{b_{mini}} + \tau}}$$
	where $\alpha_{b_{mini}}$ and $\beta_{b_{mini}}$ are the mean and variance deviation values of the input activations $B_mini$, $\varepsilon$ and $\theta$ the two parameters that can be learned. This transformation is performed at the BN layer, providing the possibility to linearly transform the normalized activation back to any scale. Since $\varepsilon$ is a parameter that scales $\breve{B}$ , we use $\varepsilon$ directly as a network pruning object. In a first step, we should perform sparse training by adding the L1-norm to $B_{out}$ as a constraint on its forward propagation during the BN layer update \cite{ref14} and calculate the parameter values $\varepsilon$ for each BN layer as a basis for judging the impact of this channel on the network. In addition, the absolute value function added by L1 regularization produces a loss function that is not differentiable everywhere, and then the weight matrix is generated as a sparse matrix by a back propagation algorithm. 
	
	\subsubsection{Pruning the weight files after training}
	For the created weight file, first find its corresponding weights according to the sequence number of the BN layer. This weight saves the value judged whether this channel is crucial. Next, the weight values will be sorted from minimum to maximum and then pruned according to the pruning ratio $\gamma$ to pruning. For example, the pruning ratio $\gamma$ is 0.7, and then the first 70\% of the weight values are set to 0 and only the deviation $\theta$ is added. When the weight information of all BN layers are estimated, the weight file is updated by the formed parameters as our training model, and a new network profile is generated. In fact, there are two pruning methods: normal pruning and regular pruning. Normal pruning implies setting the $\varepsilon$ parameter strictly to 0 according to $\gamma$ and obtaining a considerable compression ratio. Regular pruning is designed for hardware deployment and the number of filters after pruning is a multiple of 8. Regular trimming sacrifices part of the compression ratio of the application, but the mAP value of regular trimming is usually higher than regular trimming for the same compression ratio. We can clearly see the trimming process above in Figure 1. 
	\begin{figure}[htbp]
		\centering
		\includegraphics[scale=0.4]{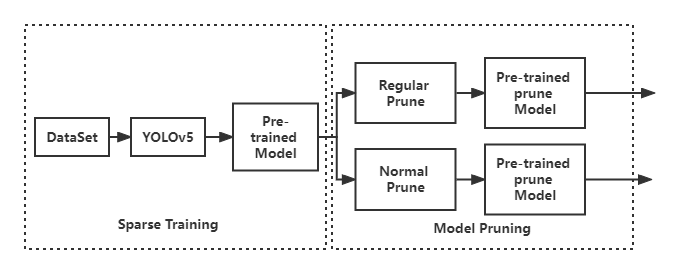}
		\caption{Fig.1.The figure depicts the process from network sparsity training to pre-training model generation and pruning model generation by two different pruning methods. }
		\label{figure}
	\end{figure}

	\section{EXPERIMENTS}
	To find the most suitable pruning method and data pre-processing method in this dataset, we will test different YOLOv5-based models and find out which data pre-processing method can have better results. 
	
	\subsection{Description of Data Set}
	The samples of the data set used in this article comes from a hundred of real roads of Chengdu. The main obstacles are vehicles and passengers, which are obtained by drone aerial photography. The size of the collected pictures is all 960 * 540, which retains the information of the object to the greatest extent.
	
	The data set has the following three characteristics:
	
	\subsubsection{Multiple Weather}The data collection selects the three local weather conditions that are more common in the local area, namely sunny, cloudy, and foggy days. Of these, 275 images were collected on sunny days, 238 images were collected on cloudy days, and 227 images were collected here on foggy days, which ensured that the total amount of data was nearly equal for all three environments.
	\subsubsection{Considerable} There were 778 valid images in this dataset, including over 1,000 vehicles, which meets the criteria for a dense dataset and the actual scenario of the road.
	\subsubsection{Comprehensive} Because of the dense vehicles and passengers. During the data collection process, the UAV was launched at three different angles: flat, pitch and elevation, which preserved the spatial information of the target to a certain extent and could solve the problem of target occlusion to a certain extent.
	
	\subsection{Performance of Different Model}
	We will first test three YOLOv5-based models on the OBSTACLE dataset. The tests are evaluated by considering both real-time and accuracy, so we compare the performance of network models generated from four perspectives: network parameter volume, model volume, time to infer individual images, and maps. We will train on a GTX1050Ti GPU and an i7 8750H CPU, always using a fixed (416*416) pixel size as input data for training, without using any other data pre-processing methods, in addition to YOLOv5 trimming using normal trimming with a trimming rate of 0.3 for training and testing. for network models generated with different trimming methods and different trimming rates The performance on the OBSTACLE dataset was also tested in detail. the performance of YOLOv5, normal pruning YOLOv5 and regular pruning YOLOv5 on the test set is shown in Tables I and II, where the performance of the two pruning methods and the network models generated with different pruning rates on the ORANGE dataset are described in detail. In the comparison experiments, we compared all pruned networks with the unpruned YOLOv5 network. By comparing the experimental results, we can find that the volume of the normally pruned model is about 8\% smaller than the volume of the conventionally pruned model after regular pruning of the network model at the same pruning rate. At the same time, the inference time of the normally pruned model for a single image is about 25\% shorter than that of the conventionally pruned model. In addition, we can find that the mapping values of the network models they generated on the orange dataset decreased to some extent, regardless of whether regular pruning or conventional pruning was used. 

	\begin{center}
	\begin{table*}[htbp]
		\center
		\caption{\centering COMPARE THE PERFORMANCE OF DIFFERENT MODELS}
		\setlength{\tabcolsep}{7mm}{
		\begin{tabular}{|c|cc|cc|}
			\hline
			\multirow{2}{*}{Model} & \multicolumn{2}{l|}{Model Processing Method}                          & \multicolumn{2}{l|}{Model Evaluation}                                      \\ \cline{2-5} 
			& \multicolumn{1}{l|}{Prune Method}  & \multicolumn{1}{l|}{Prune Ratio} & \multicolumn{1}{l|}{Model Volume} & \multicolumn{1}{l|}{Compressing Ratio} \\ \hline
			YOLO@416               & \multicolumn{1}{c|}{None}          & None                             & \multicolumn{1}{c|}{471.6MB}      & None                                   \\ \hline
			YOLO@416               & \multicolumn{1}{c|}{Normal Prune}  & 0.3                              & \multicolumn{1}{c|}{329.8MB}      & 69.9\%                                 \\ \hline
			YOLO@416               & \multicolumn{1}{c|}{Normal Prune}  & 0.5                              & \multicolumn{1}{c|}{242.3MB}      & 51.3\%                                 \\ \hline
			YOLO@416               & \multicolumn{1}{c|}{Regular Prune} & 0.3                              & \multicolumn{1}{c|}{342.5MB}      & 72.6\%                                 \\ \hline
			YOLO@416               & \multicolumn{1}{c|}{Regular Prune} & 0.5                              & \multicolumn{1}{c|}{278.6MB}      & 59.1\%                                 \\ \hline
		\end{tabular}}
	\end{table*}
	\end{center}

	\begin{center}
	\begin{table*}[htbp]
		\centering
		\caption{\centering COMPARE THE PERFORMANCE OF DIFFERENT MODELS}
		\setlength{\tabcolsep}{8mm}{
		\begin{tabular}{|c|cc|cc|}
			\hline
			\multirow{2}{*}{Model} & \multicolumn{2}{l|}{Model Processing Method}                          & \multicolumn{2}{l|}{Model Evaluation}                          \\ \cline{2-5} 
			& \multicolumn{1}{l|}{Prune Method}  & \multicolumn{1}{l|}{Prune Ratio} & \multicolumn{1}{l|}{Time Consuming} & \multicolumn{1}{l|}{mAP} \\ \hline
			YOLO@416               & \multicolumn{1}{c|}{None}          & 0                                & \multicolumn{1}{c|}{0.039}          & 75.19                    \\ \hline
			YOLO@416               & \multicolumn{1}{c|}{Normal Prune}  & 0.3                              & \multicolumn{1}{c|}{0.017}          & 73.65                    \\ \hline
			YOLO@416               & \multicolumn{1}{c|}{Normal Prune}  & 0.5                              & \multicolumn{1}{c|}{0.015}          & 72.94                    \\ \hline
			YOLO@416               & \multicolumn{1}{c|}{Regular Prune} & 0.3                              & \multicolumn{1}{c|}{0.024}          & 73.81                    \\ \hline
			YOLO@416               & \multicolumn{1}{c|}{Regular Prune} & 0.5                              & \multicolumn{1}{c|}{0.022}          & 74.79                    \\ \hline
			Mask R-CNN@416         & \multicolumn{1}{c|}{None}          & 0                                & \multicolumn{1}{c|}{0.067}          & 77.27                    \\ \hline
			Faster R-CNN@416       & \multicolumn{1}{c|}{None}          & 0                                & \multicolumn{1}{c|}{0.074}          & 77.18                    \\ \hline
		\end{tabular}}
	\end{table*}
	\end{center}
	\vspace{-1cm}
	\subsection{Experimental Results}
	In contrast, when the pruning rate y is 0.3, the mAP value of the model using regular pruning is 0.7\% lower than the mAP value of the model using normal pruning; when the pruning rate y is 0.5, the mAP value of the model using regular pruning is 2.2\% lower than the mAP value of the model using regular pruning. On the other hand, for the same pruning method, a more intuitive comparison results in the fact that as the pruning rate increases, the volume of the model generated after pruning will be smaller, the inference time will be reduced and the mapping will also be reduced. The comparison shows that when the input image size is 416*416 and the conventional pruning method is used, the model file generated with a pruning rate of 0.3 is not only compressed by 31.5\%, but also the time to infer a single image is reduced after 54\% and the mAP is only reduced by 2.4\%, which has little impact on the practical application of target detection and does not have much impact. Therefore, we selected the model file under this pruning condition, added data pre-processing methods and continued testing. 
	
	\subsection{Data Pre-processing}
	YOLOv5 is a one-stage neural network and data pre-processing plays a crucial role in the training of the network. Therefore, we will determine which method is suitable for this data by testing the effect of the following four data pre-processing methods on the YOLOv5 prune.
	
	\subsubsection{Change exposure training}
	To use this method, we set the exposure value to 1.5 and then adjust the exposure to 1.5 times the original image before feeding the training image into the network.
	\subsubsection{Random angle training}
	When using this method, we set the angle value to 5, at which point it will flip between -5° and +5° before training the image for input to the network. 
	\subsubsection{Change hue training}
	When using this method, we set the chrome value to 0.1. At this point, before feeding the training image into the network. The chrome will be adjusted to 0.1 times the original image. 
	\subsubsection{Change saturation training}
	With this method, we set the saturation value to 1.5. At this point, the saturation will be adjusted to 1.5 times the original image before the training image is fed into the network
	\subsubsection{Random shape training}
	The sample photos for this training set are taken from the scene of the drone and are all 540*960 in size. when using this method, the size of input batch size number of images will be adjusted to 
	the H set A size in, where the H set is \{320,352,384,416,448,480,512\}. 
	For the above four methods, we will use an image of size 416*416 as input and prune the training and test normally at a pruning rate of 0.3. After applying these five methods to the dataset and training with the designed network, the test results are shown in Table III. \vspace{1cm}
	
	\begin{center}
	\begin{table}[htbp]
		\caption{COMPARISON OF DIFFERENT PRE-TREATMENT METHODS }
		\setlength{\tabcolsep}{4.2mm}{
		\begin{tabular}{|c|ccc|}
			\hline
			\multirow{2}{*}{Approach}  & \multicolumn{3}{l|}{Module Evaluation}                                                \\ \cline{2-4} 
			& \multicolumn{1}{l|}{mAP}   & \multicolumn{1}{l|}{$\Delta$}     & \multicolumn{1}{l|}{Cumu $\Delta$} \\ \hline
			baseline                   & \multicolumn{1}{c|}{74.79} & \multicolumn{1}{c|}{0}     & 0                           \\ \hline
			Random Shape Training      & \multicolumn{1}{c|}{75.19} & \multicolumn{1}{c|}{+0.40} & +0.40                       \\ \hline
			Random Angel Training      & \multicolumn{1}{c|}{74.82} & \multicolumn{1}{c|}{+0.03} & +0.40                       \\ \hline
			Change Saturation Training & \multicolumn{1}{c|}{74.81} & \multicolumn{1}{c|}{+0.02} & +0.45                       \\ \hline
			Change Exposure Training   & \multicolumn{1}{c|}{74.85} & \multicolumn{1}{c|}{+0.05} & +0.49                       \\ \hline
			Change Hue Training        & \multicolumn{1}{c|}{74.83} & \multicolumn{1}{c|}{+0.04} & +0.52                       \\ \hline
		\end{tabular}}
	\end{table}
	\end{center}
	\vspace{-1cm}
	
	The comparison reveals that the network's accuracy is most pronounced when random shape training is added. When all three methods, varying saturation training, varying exposure training and varying hue training, are used simultaneously, the improvement in accuracy is not simply the effect of using them separately to superimpose the improvement in network accuracy, suggesting that the three methods have a somewhat repetitive effect in this dataset and network structure. In contrast, although random angle training and random shape training are used in combination, there is also some reduction in accuracy improvement due to the repetitive effect, but the reduction is very small and negligible. Therefore, in order to reduce the training time of the network on the basis of ensuring the accuracy improvement and keeping the original features of the dataset as much as possible, we only chose three data pre-processing methods, namely random shape training, random angle training and variation saturation training. 
	
	\section{CONCLUSION}
	In this paper we provide a obstacle dataset collected in a real-world scenario by comparing several more widely used YOLOv5 structures, comparing the model volume, inference time and number of parameters on a test set of model files generated after training on this dataset. Apart from the mapping values, YOLOv5 prune was the most suitable model for this dataset. By comparing the performance on the test set of model files generated by two different pruning methods and different pruning rates, we selected the most appropriate pruning method and pruning rate. The model files were then generated by adding different data pre-processing methods to the dataset and training using the selected pruning methods and pruning rates. Through testing, three data pre-processing methods were selected and combined to maximise network performance. All available data and training results can be replicated.

\end{document}